\documentclass[letterpaper, 10 pt, conference]{ieeeconf}

\IEEEoverridecommandlockouts
\usepackage{url}
\usepackage{amsmath}
\usepackage{amssymb}
\usepackage{algorithm}
\usepackage{algorithmic}
\usepackage{multirow}
\usepackage{graphicx}
\usepackage{booktabs}
\usepackage{subcaption}

\title{\LARGE \bf
Fleet-To-Lab: A Transfer Learning Framework For Lunar Rover Slippage Estimation Via Model Fusion}

\author{Riccardo Viviano, Saki Omi, Andrej Orsula, Miguel Olivares-Mendez%
\thanks{All authors are with the Space Robotics Research Group (SpaceR), Interdisciplinary Centre 
for Security, Reliability and Trust (SnT), University of Luxembourg, 
Luxembourg. {\tt\small \{riccardo.viviano, saki.omi, andrej.orsula, 
miguel.olivaresmendez\}@uni.lu}}}

\begin{document}

\maketitle
\thispagestyle{empty}
\pagestyle{empty}

\begin{abstract}
Accurate wheel slip estimation is essential for autonomous lunar rover mobility and navigation. Machine Learning models trained on terrestrial data generalize poorly to lunar terrain, and real lunar datasets are scarce due to the limited number of missions and costly data acquisition. We present Fleet-to-Lab, a transfer learning framework that leverages proprioceptive data collected by previously deployed heterogeneous lunar rovers to mitigate the Earth-Moon domain gap in slip estimation for a future deployable unit. We fuse
several heterogeneous expert models into a single architecture, using a modest dataset collected after the rover deployment. We propose AcoMerge, a new hybrid swarm-intelligence algorithm that performs model fusion by searching for an optimal combination of expert parameters. Experiments conducted in a high-fidelity physics simulation show balanced accuracy and macro-F1 improvements compared to deep model fusion baselines. AcoMerge exhibits competitive performance with joint training on deep architectures, while achieving higher macro-F1 and balanced accuracy on a smaller model. Overall, our framework shows model fusion as a possible transfer learning alternative for slippage estimation in space robotic missions with limited data. 
\end{abstract}

\section{INTRODUCTION}

\begin{figure*}[t]
  \centering
  \includegraphics[width=\textwidth]{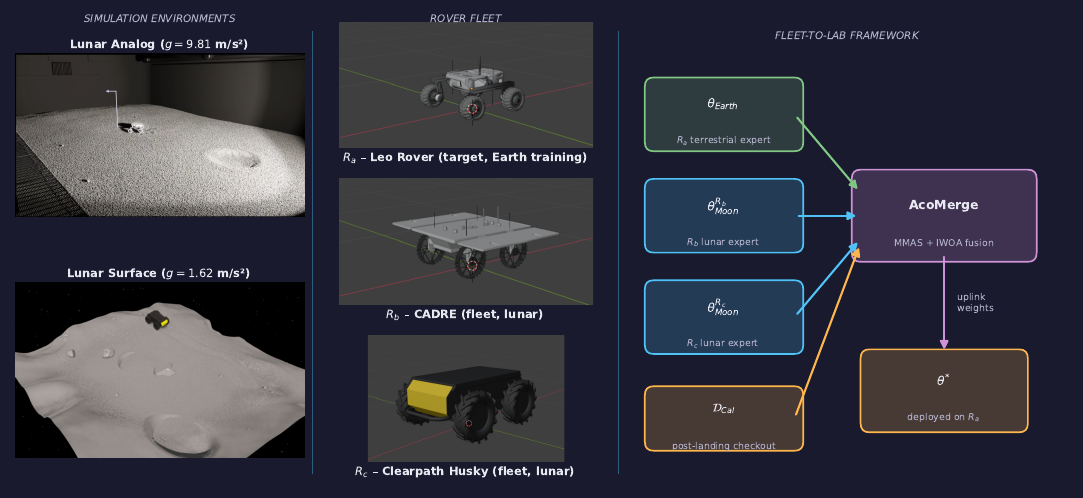}
  \caption{Fleet-to-Lab Transfer Learning framework. Fleet rovers $R_b$ and $R_c$ 
  collect proprioceptive data under lunar gravity across diverse terrain configurations. 
  Target rover $R_a$ trains exclusively in a lunar analog under terrestrial conditions (different gravity, terrain geometry, friction coefficients, particle soil properties). AcoMerge fuses all 
  three expert models guided by the post-landing calibration set $\mathcal{D}_{Cal}$, 
  producing a single deployed model $\theta^*$.}
  \label{fig:overview}
\end{figure*}

Autonomous rover mobility on the lunar surface depends critically on accurate wheel-terrain interaction estimates, especially concerning slippage. Excessive slip leads to localization errors and the catastrophic risk of entrapment~\cite{gonzalez2018slippage}. While the China Lunar Exploration Program (CLEP) has successfully deployed rovers such as Yutu (Chang'e-3) and Yutu-2 (Chang'e-4)~\cite{zuo2021china, li2015change3, li2021change4}, the wealth of telemetry generated by these missions remains a largely underutilized resource. Through initiatives such as the China Lunar and Planetary Data System (CLPDS), these in-situ datasets are now available to the research community; however, their potential for refining the mobility and navigation of future lunar vehicles has yet to be fully realized.

Proprioceptive sensing approaches for slip estimation rely on wheel odometry and inertial measurements without needing exteroceptive sensors~\cite{omura2017wheel, rw_slippage10}. The underlying problem is that slippage dynamics shift drastically depending on gravity and soil parameters, even though the underlying wheel-soil interaction dynamics remain the same. The Moon's reduced gravity fundamentally alters how wheels sink and how the soil bears weight~\cite{lopez2019do}. Because of this massive physical difference, machine learning models built in controlled Earth labs routinely fail when encountering unpredictable planetary terrain~\cite{survey_slippage_estimation}. This extreme sensitivity to domain shifts limits the reliability of standard supervised estimators~\cite{rw_slippage8, rw_slippage9}.

Transfer Learning (TL) offers a viable solution to mitigate this domain shift, as demonstrated by adapting slip predictions across varying terrain inclinations~\cite{slip_prediction_transfer_learning}. Extending on this idea, we demonstrate that data generated by actual lunar rovers from past missions can improve the performance of new vehicles trained in a terrestrial environment. Under this ``Fleet-to-New-Rover'' setup, a new rover simply inherits the neural representations learned in real lunar conditions. Feasibility is supported by archives such as The China Lunar and Planetary Data System (CLPDS)~\cite{zuo2021china}, available through the Ground Research and Application System (GRAS) portal\footnote{\url{https://moon.bao.ac.cn}}. While lacking direct slip measurements, their telemetry and orbital imagery allow retrospective ground-truth derivation: by correlating wheel odometry with visual landmarks in orbital images, precise slip values could be reconstructed. In this study, we utilize high-fidelity simulation to model these data streams, establishing a framework that can be used on processed lunar datasets as they become available.

We simulate a scenario where two rovers operate in a lunar environment while a third trains in a lunar analog with terrestrial gravity, different terrain, different friction coefficients and different particle soil properties, and we transfer learned representations from the lunar fleet to mitigate Earth--Moon domain gap degradation. All experiments are conducted in high-fidelity physics simulation.

The main contributions of this paper are:

\begin{itemize}
    \item \textbf{Fleet-to-Lab Transfer Learning Framework:} We propose a fleet-to-lab transfer framework for lunar rover slip estimation under the Earth–Moon domain gap, leveraging telemetry from prior rover platforms to improve slip estimation for a new rover.

    \item \textbf{A Hybrid Swarm Intelligence Algorithm:} We introduce AcoMerge, a hybrid swarm-intelligence model-fusion algorithm, designed to combine heterogeneous rover models into a single deployable slip estimation network enabling deployment on resource-constrained rover hardware. 

    \item \textbf{Comprehensive Experimental Evaluation:} Experiments in high-fidelity physics simulation validate the framework and AcoMerge, showing performance superior to model-merging baselines and comparable to joint fine-tuning, with statistically significant macro-F1 and balanced accuracy gains on capacity-constrained architectures suitable for onboard deployment.
\end{itemize}

\noindent\textbf{Reproducibility.} We release the data collection script, acquired datasets, \texttt{.usdc} assets for all lunar environments and rover platforms, the full training pipeline \footnote{\url{https://github.com/VivianoRiccardo/Fleet-To-Lab}}, video recordings of the rover platforms operating on the different surfaces \footnote{\url{https://drive.google.com/drive/folders/1rrHhKuD8BpEVye5IY_SY9nEWqi4pD-u_?usp=sharing}}.

\section{RELATED WORKS}

Slippage estimation for planetary rovers has been approached through classical analytical methods, data-driven
learning, and more recently transfer learning. This section
reviews each paradigm and positions our contribution relative
to existing model fusion strategies.

\subsection{Slippage Estimation in Planetary Exploration}

The literature on wheel slip estimation generally falls into three broad categories: analytical modeling, online calibration, and machine learning.

Early analytical methods heavily rely on kinematic constraints and terramechanics. For example, in~\cite{rw_slippage1}, the authors designed an Extended Kalman Filter (EKF) estimator built around a continuous tire model. Along similar lines, Reina et al.~\cite{rw_slippage2} blended proprioceptive measurements with fuzzy logic. Still, a major hurdle for these classical methods is their strict dependence on terrain-specific parameters, which are notoriously difficult to estimate when a rover is navigating unknown lunar or Martian regolith. To get around this parameter dependency, some researchers turned to online calibration techniques~\cite{rw_slippage3, rw_slippage5}. Unfortunately, these setups usually demand Real-Time Kinematic GPS (RTK-GPS), a technology unavailable in planetary environments. While there are GPS-free alternatives, like the skid-steer kinematic model proposed by~\cite{rw_slippage4}, they are generally derived from tests on Earth. Consequently, they tend to struggle when faced with the sudden domain shifts of an extraterrestrial environment.

Because of these limitations, the field has increasingly embraced machine learning to capture the highly nonlinear dynamics of wheel-soil interactions directly from raw sensor data. As demonstrated in~\cite{rw_slippage7}, learning-based estimators actually outperform classical techniques when constrained to the kinds of sensors that are practically feasible for space missions. Within this space, researchers face a trade-off between accuracy and the need for labeled data. Unsupervised frameworks, like the Gaussian Mixture Model from~\cite{rw_slippage6}, successfully detect slip events without relying on any ground truth. On the other hand, supervised methods like those in~\cite{rw_slippage9} can push accuracy beyond 96\% when label data is available.

While traditional machine learning models, such as Random Forests, create rigid decision boundaries that adapt poorly when shifted to entirely different domains~\cite{rw_slippage8}, Deep Neural Networks (DNNs) offer a way out. The capacity for representation learning makes DNNs uniquely suitable for transfer learning, as latent features learned in one domain can be adapted to another.

\subsection{Transfer Learning in Robotics}

Transfer learning deals with domain shifts by taking knowledge gained in a source domain to improve performance in a target scenario~\cite{rw_transfer_learning1}. In~\cite{rw_transfer_learning2} the authors note that transferring skills between physically different robots demands carefully aligning their internal feature representations, for example, by mapping diverse sensory data into a shared latent space or matching high-level task outcomes. Deep model fusion achieves this alignment directly in the parameter space by merging the weights of heterogeneous experts. In~\cite{rw_transfer_learning4} the authors demonstrated that successful transfer between heterogeneous robots requires robots with the same dynamics model but different parameters to avoid negative transfer.

Looking specifically at planetary exploration, in~\cite{slip_prediction_transfer_learning} a multi-source Gaussian Process Regression is used to predict slip on hazardous terrains. While this effectively handled environmental changes, it did not tackle cross-robot transfer across entirely different domains. More recently, deep learning methods, like the cross-embodiment skill transfer introduced by~\cite{rw_transfer_learning6}, have proven that finding invariant representations can successfully bridge the gap between different robot bodies. To extend these deep transfer concepts to proprioceptive sensing, we need robust strategies to merge knowledge gathered from multiple distinct source domains.

\subsection{Deep Model Fusion}

Deep model fusion aggregates multiple trained networks into a unified model ~\cite{modelfusionsurvey}. Weight averaging methods such as DARE-TIES ~\cite{dareties} are straightforward but rely on the assumption of Linear Mode Connectivity (LMC), which requires models to lie in the same loss landscape basin. This assumption breaks down under large domain gaps such as the Earth--Moon domain shift. Alignment methods \cite{adaptiveMergingTrasnferLearning} address this but require sequential training. Evolutionary and swarm-based strategies ~\cite{rw_merging3, mergingRecipes, modelSwarm} search for optimal combinations, bypassing these constraints. However, they have not been demonstrated for heterogeneous domain fusion across platforms with different dynamics, sensing, or operating environments.

\section{METHOD}

\subsection{Problem Formulation}

Our objective is to estimate the wheel slip state of a target rover $R_a$ operating in a lunar environment using only onboard proprioceptive sensing.

\subsubsection{Slip Definition}

We adopt a distance-based slip formulation defined by~\cite{gonzalez2018slippage} as: $s= 1-d_{A}/d_{C}$, where $d_A$ denotes the actual distance traveled and $d_C$ denotes the commanded distance. Following the approach defined in~\cite{gonzalez2018slippage}, we frame the problem as a discrete classification task aligned with hazard detection requirements:

\begin{equation}
y_t =
\begin{cases}
0 & \text{if } s \leq 0.3 \quad (\text{Low Slip}) \\
1 & \text{if } 0.3 < s \leq 0.6 \quad (\text{Moderate Slip}) \\
2 & \text{if } s > 0.6 \quad (\text{High Slip})
\end{cases}
\label{eq:slip_classes}
\end{equation}

\subsubsection{Learning Objective}

All rovers share a common input space $\mathbf{x}_t \in \mathbb{R}^{D}$ but differ in physical parameters and operating domains (Earth vs. Moon). We aim to learn a parametric classifier $\hat{y}_t = f_{\theta}(\mathbf{x}_t)$ that generalizes to the lunar domain.

\subsubsection{Available Data}

We assume the following datasets:

\begin{itemize}
    \item $\mathcal{D}_{Earth}^{R_a}$: terrestrial simulation data for the target rover $R_a$.
    \item $\mathcal{D}_{Moon}^{R_x}$: Historical lunar data collected from distinct rover platforms ($R_x$), where each $R_x$ represents a unique vehicle.
    \item $\mathcal{D}_{Cal}$: a limited calibration dataset collected by $R_a$ after lunar landing.
\end{itemize}

Each dataset is $\mathcal{D} = \{(\mathbf{x}_t, y_t)\}_{t=1}^{N}$, where $y_t$ is obtained from Eq.~\ref{eq:slip_classes}. Let $\theta_{Earth}$ denote parameters trained on $\mathcal{D}_{Earth}^{R_a}$ and $\theta_{Fleet} = \{\theta_{Moon}^{R_b}, \theta_{Moon}^{R_c}\}$ those trained on lunar fleet data. The objective is:

\begin{equation}
\theta^{*} = \Phi(\theta_{Earth}, \theta_{Fleet} \mid \mathcal{D}_{Cal})
\end{equation}

that minimizes target lunar-domain classification error.

\subsubsection{Calibration Dataset}
\label{sec:calibration}

The calibration dataset $\mathcal{D}_{Cal}$ is a small corpus of labeled proprioceptive data that $R_a$ obtains during an initial post-landing checkout traverse. Ground-truth slip labels can be obtained either onboard, by correlating camera imagery with commanded trajectories via visual odometry~\cite{visualOdometry1,visualOdometry2} and manual review, or retrospectively, by matching telemetry against orbital imagery to reconstruct the path traveled. Being both approaches labor-intensive due to, either the need of human verification in the loop, or the adoption of challenging methods~\cite{Chen2024} for mapping ground views and orbital images, we use a small collected checkout data as calibration dataset. Since even a complete autonomous drive spans only a few hundred meters between position fixes~\cite{LunarNav}, a checkout covers correspondingly less. $\mathcal{D}_{Cal}$ is restricted to a number of samples of size $|{D}_{Cal}| \in \left \{150,200 \right \}$, at the 2s stride, $\sim$5-7 min of driving and $\sim$45-60m of traversed terrain, drawn from 8 manual traverses, with an approximate class distribution of $65\%$ low, $22 \%$ moderate, $13 \% $ high slip.

\subsection{State Space and Feature Extraction}
\label{sec:features}

Based on the windowing strategies in~\cite{rw_slippage9}, we apply the raw proprioceptive signals using a sliding temporal window $W_{t}$ of 20 timesteps (2 seconds at 10 Hz), with 2 seconds stride so successive samples share no raw signal, condensed into a feature vector $\mathbf{x}_{t} \in \mathbb{R}^{D}$ 
by calculating the mean ($\mu$) and standard deviation ($\sigma$) for each channel:

\begin{equation}
\mathbf{x}_t = \left[ \mathbf{f}_{prop},\; \mathbf{f}_{imu},\; \mathbf{f}_{att} \right]^T
\end{equation}

Where $\mathbf{f}_{prop}$ contains the motor current and wheel velocity statistics ($[\mu_{I}, \sigma_{I}, \mu_{\omega_w}, \sigma_{\omega_w}]$), and $\mathbf{f}_{imu} = [\mu_{a_x}, \sigma_{a_x}, \mu_{a_y}, \sigma_{a_y}, \mu_{a_z}, \sigma_{a_z}]$ covers linear acceleration across all axes. We include only the mean roll and pitch ($[\mu_{roll}, \mu_{pitch}]$) for the attitude vector $\mathbf{f}_{att}$ since load oscillations and vibration severity are already reflected in the dispersion of current and acceleration signals~\cite{gonzalez2018slippage}. We fed models with raw data without any normalization since it degraded transfer in our preliminary experiments.

\subsection{AcoMerge: Graph-Based Deep Model Fusion}
\label{sec:xacomerge}

Since all rovers share the same feature dimensionality $D$, every model uses an identical DNN architecture parameterized by $\theta$. We treat the fusion of source models ($\theta_{Earth}$, $\theta_{Fleet}$) into $\theta^{*}$ as a meta-heuristic search problem. AcoMerge decomposes this into two phases: Phase~1 solves a combinatorial problem (selecting the best expert source per parameter); Phase~2 solves a continuous problem (interpolating between expert values to find optimal weights not corresponding to any single source).

\begin{figure}[t]
  \centering
  \includegraphics[width=\columnwidth]{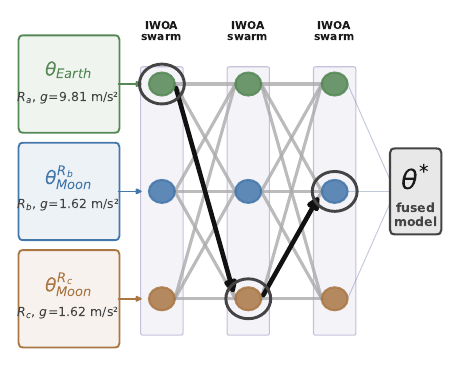}
  \caption{Fusion graph $\mathcal{G}$. Each node column (purple) is an independent IWOA swarm. The ant path selects one expert per parameter; circled nodes are the chosen candidates.}
  \label{fig:fusion_graph}
\end{figure}

\begin{figure}[t]
  \centering
  \includegraphics[width=\columnwidth]{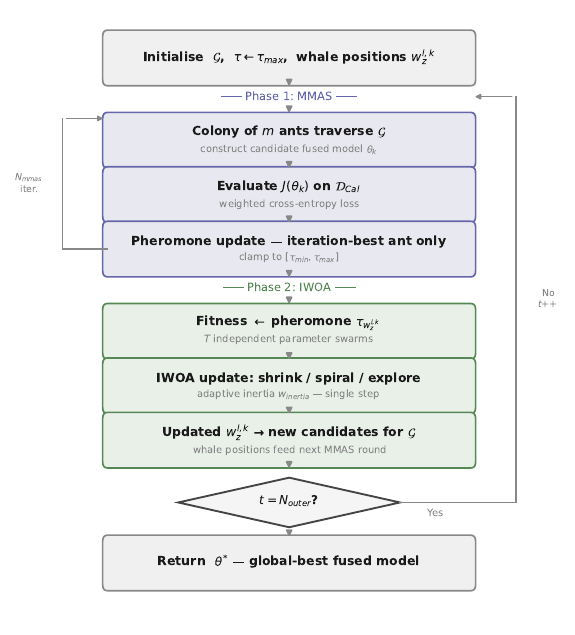}
  \caption{AcoMerge algorithm. Phase~1 (MMAS) and Phase~2 (IWOA) alternate every $N_{mmas}$ iterations over $N_{outer}$ outer loops, with pheromone signals tying the two phases together.}
  \label{fig:flowchart}
\end{figure}

\subsubsection{Fusion Graph Topology}

Fusion is represented as a path-finding problem on a directed acyclic graph $\mathcal{G} = (\mathcal{V}, \mathcal{E})$. For a fully connected architecture with $L$ layers, the forward pass of layer $l$ is given by:

\begin{equation}
\label{eq:feedforward}
\mathbf{a}_{l} = \sigma_l(\mathbf{W}_l \mathbf{x}_{l} + \mathbf{b}_l)
\end{equation}

Given $K$ pre-trained experts, the weight matrix $\mathbf{W}_l^k$ and bias $\mathbf{b}_l^k$ of each expert $k$ are flattened into a vector $\mathbf{v}_l^k$ of size $Z_l$. Each scalar $w_{z}^{l,k}$ is a candidate node, yielding $K$ parallel candidates at each index $z \in [0, Z_l-1]$. Edges connect $w_{z}^{l,k}$ to every candidate $w_{z+1}^{l,v}$ for all $k, v \in [0, K-1]$ (intra-layer), and the final nodes of layer $l$ connect to the initial nodes of layer $l+1$ (layer termination). An ant traversing $\mathcal{G}$ constructs a candidate fused model by selecting exactly one expert source per learnable parameter (Fig.~\ref{fig:fusion_graph}).

\subsection{Phase 1: Constructive Search via Max-Min Ant System}
\label{sec:phase1}

We use the MMAS variant~\cite{acs}, where the pheromone range is bounded in $[\tau_{min}, \tau_{max}]$ limits early convergence, especially as the search space contains $K^{T}$ configurations, where $T = \sum_{l=0}^{L-1} Z_l$. A colony of $m$ ants constructs candidate fusion models by traversing $\mathcal{G}$ with a pseudo-random-proportional transition rule:

\begin{equation}
\label{eq:aco_transition}
p^{k}_{s_{t},s_{t+1}}(n) =
\begin{cases}
    1, & \text{if } \rho \leq \rho_{0} \text{ and}\\
    & \tau_{s_{t+1}} = \max_{s \in S_{s_{t}}^{k}} \tau_{s}\\[1ex]
    \dfrac{\tau_{s_{t+1}}(n)^{\alpha}\;\eta_{s_{t},s_{t+1}}^{\beta}}
    {\displaystyle\sum_{s \in S_{s_{t}}^{k}} \tau_{s}(n)^{\alpha}\;\eta_{s_{t},s}^{\beta}}, 
    & \text{if } \rho > \rho_{0}\\[1ex]
    0, & \text{otherwise}
\end{cases}
\end{equation}

where $\tau_{s_{t+1}}(n)$ is pheromone on node $s_{t+1}$, $\eta_{s_{t},s_{t+1}}$ is the heuristic (set to the constant value of 1 in our experiments), $\alpha$, $\beta$ control their relative influence, and $\rho_0 \in [0,1]$ governs exploitation-exploration balance. Each ant evaluates $\theta_k$ on $\mathcal{D}_{Cal}$ via weighted cross-entropy $J(\theta_k)$ and deposits pheromone $\Delta\tau_{s_{t+1}}^{k}(n) = Q / J(\theta_k)$ on visited nodes, where $Q$ is a constant. Only the iteration-best ant contributes to the global update:

\begin{equation}
\label{eq:pheromone_update}
\tau_{s}(n+1) = \rho_{evap}\;\tau_{s}(n) + (1-\rho_{evap})\;\Delta \tau_{s}^{best}(n)
\end{equation}

Pheromone is clamped to $[\tau_{min}, \tau_{max}]$, where $\tau_{min}$ is set as a function of $\tau_{max}$, $\rho_{dec}$, and the mean branching factor following standard MMAS practice~\cite{acs}. A global-best pheromone update is applied periodically every $n_{gb}$ iterations to reinforce the overall best solution.

\subsection{Phase 2: Refinement via Improved Whale Optimization Algorithm}
\label{sec:phase2}

While Phase~1 identifies which expert parameter to select, optimal weights may lie \textit{between} expert values. Phase~2 treats each parameter index as an independent continuous optimization subproblem via IWOA~\cite{ArticleLiu2023Improved}. The phases are \textit{interleaved}: after every $N_{mmas}$ MMAS iterations, a single IWOA update is applied to all parameter swarms (Fig.~\ref{fig:flowchart}). This couples pheromone-guided refinement with updated whale positions over $N_{outer}$ iterations.

For each parameter index $z$ within layer $l$, we treat the $K$ candidate values $\{w_z^{l,0}, \dots, w_z^{l,K-1}\}$ as a distinct swarm of $K$ whales, resulting in $T$ independent swarms. The fitness of an individual whale is derived from the pheromone $\tau_{w_z^{l,k}}$ recorded during preceding MMAS iterations. 

IWOA was selected over Particle Swarm Optimization based on superior calibration 
set performance.

A uniform probability distribution $p \in [0,1]$ determines whether each whale follows the shrinking encircling mechanism or the spiral position update.

\textbf{Shrinking Encircling.} If $p < 0.5$ and $|A| < 1$ (exploitation):

\begin{equation}
\label{eq:iwoa_shrink}
w_z^{l,k}(t+1) = w_{z,best}^{l,t} - w_{inertia} \cdot A \cdot D
\end{equation}

where $A = \alpha(2 \cdot rand - 1)$, $D = | C \cdot w_{z,best}^{l,t} - w_z^{l,k}(t) |$, $C = 2 \cdot rand$, $\alpha$ linearly decreases from 2 to 0, and $w_{inertia} = 1 - 2\left(t/N_{outer}\right)^{3}$

\textbf{Spiral Update.} If $p \geq 0.5$:

\begin{equation}
\label{eq:iwoa_spiral}
w_z^{l,k}(t+1) = w_{inertia} \cdot D' \cdot e^{b\lambda} \cos(2\pi \lambda) + w_{z,best}^{l,t}
\end{equation}

where $D' = |w_{z,best}^{l,t} - w_z^{l,k}(t)|$, $b$ is a shape constant, and $\lambda \in [-1, 1]$ is uniform random.

\textbf{Exploration.} If $p < 0.5$ and $|A| \geq 1$, the whale moves toward a randomly selected swarm member instead of the global best, promoting diversity. After each IWOA update, modified whale positions become the new candidate values for the next MMAS round.

\section{EXPERIMENTAL EVALUATION}

\subsection{Simulation Platform and Rover Fleet}

All experiments use the Space Robotics Bench~\cite{orsula2025spaceroboticsbenchrobot}, built on NVIDIA Isaac Sim for high-fidelity rigid-body dynamics, deformable terrain interaction, and configurable gravity. The rover fleet consists of three heterogeneous platforms: \textbf{$R_a$ (Leo Rover)}, a four-wheeled skid-steer target vehicle trained in terrestrial conditions and evaluated under lunar gravity; \textbf{$R_b$ (CADRE)}, a lightweight JPL-based lunar rover; and \textbf{$R_c$ (Clearpath Husky)}. The rovers are released in .usdc format along with the training code.

\subsection{Data Collection and Environment Design}

Data was collected by driving each rover manually across all terrain configurations yielding a total of $\approx$ 9000 samples across 4 datasets. Fleet rovers collect data across 12 distinct lunar surface configurations parameterized by different terrain geometry (3 different lunar surfaces released in .usdc format) and friction coefficients (static friction $\in [0.90,1]$ and dynamic friction $\in [0.86, 0.94]$) under $g = 1.62\;\text{m/s}^2$; $R_a$ collects across 5 simulation seeds from the simulated lunar analog provided by default by Space Robotics Bench, with an alternative set of friction parameters,  under $g = 9.81\;\text{m/s}^2$. The lunar and moon analog terrains have been parameterized with distinct particle properties (cohesion $ \in [0.01, 0.15]$, density $\in [1500,1600]$, friction $\in [0.9,1]$, adhesion $\in [0.01, 0.05] $). Without claiming the reproducibility of lunar regolith composition, this setup has been defined to show that our approach remains effective under this kind of soil-parameter shift.

Raw data reflect the mission's preference for safer, low-slip paths, yielding a natural imbalance toward low-slip samples and relatively few high-slip datapoints. We retain these raw distributions without rebalancing, handling the imbalance through balanced accuracy, macro-F1, and weighted cross-entropy rather than resampling.

\subsection{Network Architectures}
In our experiments we consider two different MLP models.

\textbf{SlippageNet-Shallow (SN-S):} Two layers with a single 30-neuron hidden layer a normalization layer, dropout ($p = 0.2$), GELU activation; $T_{param} = 543$ parameters.

\textbf{SlippageNet-Deep (SN-D):} Four layers with hidden dimensions $[128, 64, 32]$, LayerNorm after each hidden layer, GELU activations, dropout ($p = 0.2$) after the first hidden layer; $T_{param} = 13{,}301$ parameters.

Both receive $D = 12$ input features and output logits over 3 slip classes. 
The two architectures serve complementary roles in our evaluation. SN-S is intentionally small: its parameter count puts it in line with prior proprioceptive slip estimation works~\cite{rw_slippage8}, and its small size ($T = 543$) makes it suitable for onboard planetary rover hardware with limited computational resources. SN-D is included to assess whether the fusion benefits observed on the smaller model hold true for larger networks, effectively testing if AcoMerge scales well outside of resource-constrained environments.

\subsection{Baselines}

\begin{table*}[t]
\caption{Classification performance on the lunar test set. Mean $\pm$ std over 30 MCCV runs. Best in bold; second best underlined. $\dagger$ denotes statistically significant improvement of best result over second one ($p < 0.05$, corrected resampled t-test [Nadeau $\&$ Bengio], which accounts for train/test overlap across MCCV folds).}
\label{tab:results}
\centering
\setlength{\tabcolsep}{8pt}
\begin{tabular}{ll cc cc}
\toprule
\multirow{2}{*}{\textbf{Group}} & \multirow{2}{*}{\textbf{Method}}
    & \multicolumn{2}{c}{$|\mathcal{D}_{Cal}|=150$}
    & \multicolumn{2}{c}{$|\mathcal{D}_{Cal}|=200$} \\
\cmidrule(lr){3-4} \cmidrule(lr){5-6} 
    & & Bal.\ Acc. & Macro F1
    & Bal.\ Acc. & Macro F1 \\
\midrule
\multicolumn{6}{c}{\textbf{SlippageNet-Deep (SN-D) - (MLP, 13,301 parameters)}} \\
\midrule
\multirow{3}{*}{Fine-tuned} & Joint + FullFinetune & $\mathbf{0.6617 \pm 0.0239}$ & $\mathbf{0.6312 \pm 0.0254}$ & $\mathbf{0.6803 \pm 0.0221}$ & $\mathbf{0.6515 \pm 0.0222}$ \\
& Earth + FullFinetune & $0.6211 \pm 0.0336$ & $0.5824 \pm 0.0409$ & $0.6497 \pm 0.0229$ & $0.6122 \pm 0.0302$ \\
& Model Base           & $0.4966 \pm 0.0401$ & $0.3999 \pm 0.0756$ & $0.5287 \pm 0.0473$ & $0.4465 \pm 0.0853$ \\
\midrule
\multirow{4}{*}{Merging} & Adamerging~\cite{yang2024adamergingadaptivemodelmerging}
    & $0.4859 \pm 0.0264$ & $0.4489 \pm 0.0430$ & $0.4978 \pm 0.0223$ & $0.4719 \pm 0.0287$ \\
& Evolutionary Merge~\cite{mergingRecipes} 
    & $0.4608 \pm 0.0341$ & $0.4308 \pm 0.0544$ & $0.4650 \pm 0.0272$ & $0.4367 \pm 0.0411$ \\
& ModelSwarm~\cite{modelSwarm}
    & $0.4647 \pm 0.0452$ & $0.4105 \pm 0.0442$ & $0.4739 \pm 0.0390$ & $0.4205 \pm 0.0389$ \\
& \textbf{AcoMerge (Ours)} & $\underline{0.6434 \pm 0.0237}$ & $\underline{0.6162 \pm 0.0287}$ & $\underline{0.6637 \pm 0.0270}$ & $\underline{0.6328 \pm 0.0292}$ 
\\
\midrule
\multirow{1}{*}{Distillation} & Ensemble 
distillation~\cite{hinton2015distillingknowledgeneuralnetwork}
    & $0.4669 \pm 0.0560$ & $0.4384 \pm 0.0597$ & $0.4720 \pm 0.0377$ & $0.4493 \pm 0.0381$ \\
\midrule
\multicolumn{6}{c}{\textbf{SlippageNet-Shallow (SN-S) - (MLP, 543 parameters)}} \\
\midrule
\multirow{3}{*}{Fine-tuned} & Joint + FullFinetune & $\underline{0.6202 \pm 0.0177}$ & $0.5771 \pm 0.0264$ & $0.6284 \pm 0.0173$ & $0.5890 \pm 0.0245$ \\
& Earth + FullFinetune & $0.6201 \pm 0.0218$ & \underline{$0.5854 \pm 0.0331$} & \underline{$0.6470 \pm 0.0180$} & $\underline{0.6188 \pm 0.0215}$ \\
& Model Base
    & $0.4824 \pm 0.0347$ & $0.3550 \pm 0.0639$
    & $0.4931 \pm 0.0347$ & $0.3711 \pm 0.0601$ \\
\midrule
\multirow{4}{*}{Merging} & Adamerging~\cite{yang2024adamergingadaptivemodelmerging}
    & $0.4153 \pm 0.0216$ & $0.3281 \pm 0.0492$ & $0.4172 \pm 0.0178$ & $0.3328 \pm 0.0384$ \\
& Evolutionary Merge~\cite{mergingRecipes}
    & $0.5297 \pm 0.0174$ & $0.5254 \pm 0.0202$ & $0.5256 \pm 0.0219$ & $0.5206 \pm 0.0312$ \\
& ModelSwarm~\cite{modelSwarm}
    & $0.5384 \pm 0.0163$ & $0.5091 \pm 0.0215$ & $0.5385 \pm 0.0157$ & $0.5069 \pm 0.0179$ \\
& \textbf{AcoMerge (Ours)} & $\mathbf{0.6535 \pm 0.0187}^{\dagger}$ & $\mathbf{0.6250 \pm 0.0253}^{\dagger}$ & $\mathbf{0.6732 \pm 0.0166}^{\dagger}$ & $\mathbf{0.6449 \pm 0.0228}^{\dagger}$ \\
\midrule
\multirow{1}{*}{Distillation} & Ensemble 
distillation~\cite{hinton2015distillingknowledgeneuralnetwork}
    & $0.3788 \pm 0.0518$ & $0.3420 \pm 0.0806$ & $0.3997 \pm 0.0535$ & $0.3712 \pm 0.0772$ \\
\bottomrule
\end{tabular}
\end{table*}

\textbf{Fine-Tuned Baselines:} \textit{Joint + FullFinetune} trains on $\mathcal{D}_{Earth}^{R_a} \cup \mathcal{D}_{Moon}^{R_b} \cup \mathcal{D}_{Moon}^{R_c}$ then fine-tunes all weights on $\mathcal{D}_{Cal}$, the adaptation baseline requiring full fleet data. \textit{Earth + FullFinetune} trains exclusively on $\mathcal{D}_{Earth}^{R_a}$ then fine-tunes on $\mathcal{D}_{Cal}$. This allows us to measure the effect of domain adaptation alone, without the influence of lunar data. \textit{Model Base} trains from scratch on $\mathcal{D}_{Cal}$ only, establishing the lower bound without knowledge transfer.

\textbf{Model Merging Baselines:} All operate on the same three independently trained expert models, pre-aligned using Git Re-Basin as a weight-matching algorithm~\cite{gitrebasin} to resolve permutation symmetry. \textit{ModelSwarm}~\cite{modelSwarm} applies PSO to collaboratively search model weight space. \textit{AdaMerging}~\cite{yang2024adamergingadaptivemodelmerging} learns per-tensor coefficients to combine expert task vectors, optimized here on $\mathcal{D}_{Cal}$ via weighted cross-entropy for parity with the other supervised baselines. Finally, \textit{Evolutionary Model Merge}~\cite{mergingRecipes} uses CMA-ES to optimize layer-wise DARE-TIES merging coefficients.

\textbf{Ensemble Distillation:} To explicitly test whether simpler 
output-space supervision can replace parameter search, we implement multi-teacher 
Knowledge Distillation~\cite{hinton2015distillingknowledgeneuralnetwork}. Using the three  experts as frozen teachers, a fresh student $\theta_{S}$ is trained on $\mathcal{D}_{Cal}$  by minimizing a convex combination of a temperature-scaled KL term and a weighted cross-entropy.

AcoMerge and ModelSwarm both operate over an expert buffer of 48 models: the three base experts plus 15 Gaussian-perturbed copies per expert ($\sigma_{aug} = 0.05$), expanding the combinatorial search space and improving robustness to local optima.
Buffer augmentation is not applied to Evolutionary Merge or Adamerging
due to architectural incompatibility with their respective optimization 
schemes.

\subsection{Experimental Setup and Metrics}

Performance is evaluated on the held-out test set from $R_a$ under lunar gravity in unseen lunar terrains. For each combination of method, architecture, and calibration size $|\mathcal{D}_{Cal}| \in \{150, 200\}$, we perform 30 independent Monte Carlo Cross-Validation (MCCV) runs, randomly sampling a new calibration subset each run and reporting mean $\pm$ std. We use balanced accuracy (average per-class recall) and macro-F1 score, the latter ensuring the safety-critical High Slip class contributes equally despite its low frequency. 

\subsection{Implementation and Computational Efficiency}
All gradient-based methods are tuned by grid search on the calibration set over 10 learning rates in $[10^{-4}, 10^{-1}]$ and batch sizes $B \in \{5\%, 10\%, 25\%, 50\%, 75\%, 100\%\}$ of $|\mathcal{D}_{Cal}|$; while for the merging baselines we used the hyperparameters from the original papers~\cite{dareties, mergingRecipes, modelSwarm}. We noticed AcoMerge being most sensitive to $\rho_{evap}$ set to 0.5 and $N_{ants}$ set to 300; all other hyperparameters produced \( < \) 1\% macro F1 variation across a grid sweep over the calibration set, indicating robustness to their exact values. To ensure a fair computational budget, every method is allocated exactly 130 optimization steps, one pheromone update for AcoMerge and one gradient update for Adam-based methods. Although AcoMerge evaluates $N_{ants}$ candidate solutions per step, 
all evaluations are independent forward passes admitting full 
parallelization; on an NVIDIA RTX 5060 Laptop GPU, one fully 
parallelized pheromone update requires $9.5 \pm 2.95$\,ms versus 
$9.0 \pm 1.0$\,ms for one Adam step ($1.0\times$ wall-clock ratio).

\subsection{Main Results}
\label{sec:main_results}

\begin{figure*}[t]
  \centering
  \includegraphics[width=\textwidth]{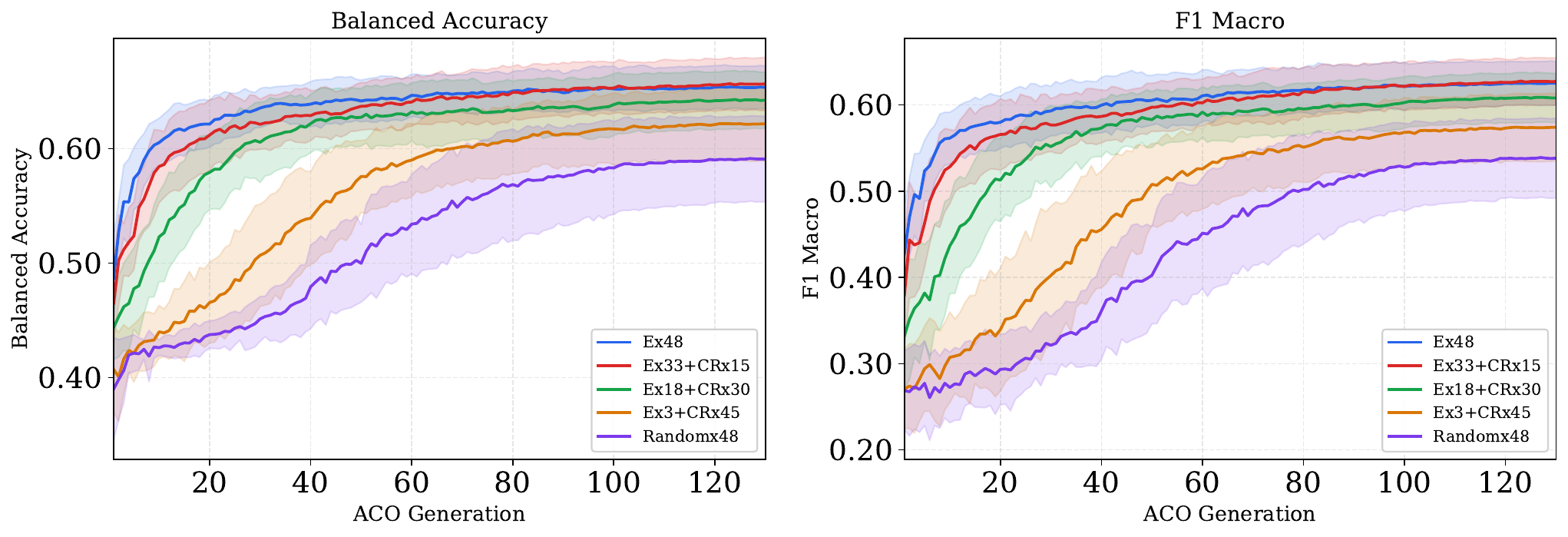}
  \caption{Effect of expert buffer composition on fusion performance (SN-S, 
$|\mathcal{D}_{Cal}|=150$, 30 MCCV runs). Balanced accuracy (left) and 
macro F1 (right) over the test samples; shaded bands denote $\pm 1\sigma$ 
across runs.}
  \label{fig:buffer_ablation}
\end{figure*}

Table~\ref{tab:results} reports results for SN-D and SN-S. On \textbf{SN-D}, Joint+FullFinetune achieves the highest balanced accuracy and macro F1 at both budgets, with AcoMerge a close second; a corrected resampled t-test on 30 paired MCCV scores confirms no statistically significant difference ($p > 0.05$), indicating AcoMerge matches joint training without requiring a combined training pipeline. On \textbf{SN-S}, AcoMerge achieves the highest balanced accuracy and Macro F1 at both budgets with statistically significant improvements over the second best ($p < 0.05$). Joint+FullFinetune shows a positive transfer from the moon fleet on \textbf{SN-D} with a statistical significance improvement ($p < 0.05$) when compared to Earth+FullFinetune, confirming that lunar fleet data provides signal terrestrial training cannot replicate, while on \textbf{SN-S} the 2 models do not outperform each other under the corrected resampled t-test. This reversal could be explained by a capacity bottleneck: with only $T=543$ parameters, a jointly trained SN-S must compress three heterogeneous domains into a single weight vector, forcing a compromise representation that loses domain-specific discriminative structure. AcoMerge instead preserves independently trained specialists and fuses them without reconciling conflicting domain statistics. On SN-D, the larger capacity alleviates this compression pressure; verifying this hypothesis remains future work. Model Base performs worst across all conditions. All merging baselines underperform the adaptation methods as well as AcoMerge, while the ensemble distillation has the worst outcome. Ablating either phase alone (MMAS, or IWOA via ModelSwarm's search with PSO replaced) kept balanced accuracy $<0.6$ and macro-F1 $<0.56$ across all settings, confirming the hybrid is needed. Regarding the impact of the calibration budget, comparing $|D_{Cal}| = 150 $ with $200$ demonstrates that while all methods improve with additional target-domain data, the relative performance rankings remain unchanged. This confirms that modestly increasing post-landing data with these constrained budgets cannot overcome the Earth–Moon domain gap alone, highlighting the necessity of transferring prior fleet knowledge.

\subsection{Effect of Expert Buffer Augmentation}
\label{sec:ablation_buffer}

The expert buffer expands three base experts with Gaussian-perturbed copies, providing a richer combinatorial search space. To determine whether gains stem from proximity to trained solutions or mere parameter space coverage, we compare balanced accuracy and F1-Macro results of five buffer configurations of identical size (48 models), progressively replacing expert-perturbed candidates ($\sigma = 0.05$) with centroid-random or fully random models.

As visible in Figure~\ref{fig:buffer_ablation} performance degrades monotonically as the proportion of expert-perturbed candidates decreases, with the fully random buffer performing worst. The gap between \textbf{E$\times$48} and \textbf{E$\times$3+CR$\times$45} isolates the contribution of local proximity: both retain the three base experts and cover comparable parameter space volume, yet removing perturbed copies in favor of centroid-random candidates causes a consistent drop. This confirms that neighborhood structure around trained solutions, not spatial coverage alone, enables effective combinatorial search. 

\subsection{Per-Class Performance Analysis}

For resource-constrained planetary deployments, SN-S is the most viable onboard architecture. To assess operational safety beyond aggregate metrics, we analyze per-class precision and recall. In autonomous rover exploration, the  trade-off depends on mission risk: higher recall on severe slip reduces the risk of wheel entrapment, while higher precision avoids safety halts that degrade traverse efficiency. On the safety-critical High Slip class, AcoMerge achieves the best operational balance, with the highest precision ($0.55 \pm 0.05$) and recall ($0.70 \pm 0.05$), while Joint+FullFinetune shows a high false alarm rate (precision $0.47 \pm 0.05$). In the transitional moderate slip regime, a key early warning indicator for soil sinkage, AcoMerge also improves sensitivity, reaching a recall of $0.55 \pm 0.04$ versus $0.46 \pm 0.09$ and $0.49 \pm 0.06$ for terrestrial and joint training, respectively. These results suggest that for capacity-bottlenecked networks, combinatorial parameter search yields more favorable precision-recall tradeoff than joint training.

\section{CONCLUSIONS}

We presented Fleet-to-Lab, a transfer learning framework for lunar rover wheel slip estimation that bridges the Earth–Moon domain gap by transferring operational knowledge from heterogeneous fleets to a new vehicle. Inheriting historical planetary experience improves slip estimation over purely terrestrial training with only a modest post-landing calibration dataset. Our proposed AcoMerge, a hybrid swarm-intelligence algorithm fusing expert models in parameter space, outperforms merging baselines and joint training on resource-constrained networks with statistically significant gains in macro-F1 and balanced accuracy, establishing model fusion as a viable pathway to repurpose scarce archival telemetry for future missions. Future work will target physical testbed validation, retrospective application to real mission datasets, and robustness to noisy labels, an important consideration given visual odometry measurement noise.

\addtolength{\textheight}{-1cm} 

\section*{ACKNOWLEDGMENT}

This research was funded in whole, or in part, by the Luxembourg National Research Fund (FNR), grant reference C24/IS/18990533/NeverGiveUp. For the purpose of open access, and in fulfilment of the obligations arising from the grant agreement, the authors have applied a Creative Commons Attribution 4.0 International (CC BY 4.0) license to any Author Accepted Manuscript version arising from this submission.

\bibliography{references}

\end{document}